# A Multimodal Feature Distillation With Mamba-Transformer Network for Brain Tumor Segmentation With Incomplete Modalities

Ming Kang, *Member, IEEE*, Fung Fung Ting, *Member, IEEE*, Shier Nee Saw, *Member, IEEE*, Raphaël C.-W. Phan, *Senior Member, IEEE*, Zongyuan Ge, *Senior Member, IEEE*, and Chee-Ming Ting, *Senior Member, IEEE*

***Abstract*—Existing brain tumor segmentation methods usually utilize multiple magnetic resonance imaging (MRI) modalities in brain tumor images for segmentation, which can achieve better segmentation performance. However, in clinical applications, some modalities are often missing due to resource constraints, resulting in significant performance degradation for methods that rely on complete modality segmentation. In this article, we propose a multimodal feature distillation with Mamba-Transformer hybrid network (MMTSeg) for accurate brain tumor segmentation with missing modalities. We first employ a multimodal feature distillation (MFD) module to distill feature-level multimodal knowledge into different unimodalities to extract complete modality information. We further develop an unimodal feature enhancement (UFE) module to model the semantic relationship between global and local information. Finally, we built a cross-modal fusion (CMF) module to explicitly align the global correlations across modalities, even when some modalities are missing. Complementary features within and across modalities are refined by the Mamba-Transformer hybrid architectures in both the UFE and CMF modules, dynamically capturing long-range dependencies and global semantic information for complex spatial contexts. A boundary-wise loss function is employed as the segmentation loss of the proposed MMTSeg to minimize boundary discrepancies for a distance-based metric. Our ablation study demonstrates the importance of the proposed feature enhancement and fusion modules in the proposed network and the Transformer with Mamba block for improving the performance of brain tumor segmentation with missing modalities. Extensive experiments on the BraTS 2018 and BraTS 2020 datasets demonstrate that the proposed MMTSeg framework outperforms state-of-the-art methods when modalities are missing.**

***Impact Statement*—Brain tumor segmentation with missing modalities is an active research area focusing on maintaining robust segmentation performance despite incomplete imaging data, which refers to techniques and approaches used in medical imaging when certain types of imaging data (modalities) are unavailable or incomplete. In brain tumor analysis, medical professionals use MRI sequences to identify and segment tumors. However, in real-world scenarios, not all these modalities are always available due to limitations like cost, time, or patient conditions. The hybrid Mamba-Transformer approach introduced in this article equips the Transformer architecture with the Mamba block as its input layer, combining both sequential and spatial information for unimodal refinement and cross-modal synthesis. MMTSeg is facilitated by long-range spatial dependencies, which force the model to learn from both available and missing modalities via the Mamba-Transformer hybrid network. With a significant increase in segmentation accuracy under incomplete modalities, the proposed MMTSeg outperforms most advanced methods.**

***Index Terms*—Cross-modal fusion (CMF), feature self-distillation, missing modalities, multimodal MRI, state space model (SSM).**

Received 1 September 2025; revised 20 November 2025 and 23 January 2026; accepted 19 March 2026. Date of publication 30 March 2026; date of current version 31 August 2026. This work was supported in part by Monash University Malaysia, the University of Malaya, and the Ministry of Science, Technology and Innovation (MOSTI) under the Malaysia-China Strategic Research Fund under Grant SRF01251347. This article was recommended for publication by Associate Editor Guilherme Nelson DeSouza upon evaluation of the reviewers' comments. *(Corresponding author: Chee-Ming Ting.)*

Ming Kang is with the School of Information Technology, Monash University, Subang Jaya 47500, Malaysia, also with the Augmented Intelligence and Multimodal Analytics (AIM) for Health Lab, Faculty of Information Technology, Monash University, Clayton, VIC 3800, Australia, and also with Monash Medical AI Group, Monash University, Clayton, VIC 3800, Australia (e-mail: ming.kang@monash.edu).

Fung Fung Ting, Raphaël C.-W. Phan, and Chee-Ming Ting are with the School of Information Technology, Monash University, Subang Jaya 47500, Malaysia (e-mail: ting.fung@monash.edu; raphael.phan@monash.edu; ting.cheeming@monash.edu).

Shier Nee Saw is with the Department of Artificial Intelligence, Faculty of Computer Science and Information Technology, Universiti Malaya, Kuala Lumpur 50603, Malaysia (e-mail: sawsn@um.edu.my).

Zongyuan Ge is with the Augmented Intelligence and Multimodal Analytics (AIM) for Health Lab, Faculty of Information Technology, Monash University, Clayton, VIC 3800, Australia, and also with Monash Medical AI Group, Monash University, Clayton, VIC 3800, Australia (e-mail: zongyuan.ge@monash.edu).

## I. Introduction

Accurate segmentation of brain tumors is crucial for tumor diagnosis and treatment. Multimodal image segmentation has emerged in medical imaging processing [1]. Multimodal magnetic resonance imaging (MRI) images exhibit various imaging features, including fluid-attenuated inversion recovery (FLAIR), T1-weighted (T1), enhanced T1-weighted (T1ce), and T2-weighted (T2) sequences, as illustrated in Fig. 1. Different modalities typically contain varying information, so multimodal complementary information can be used to enhance the model's feature representation ability. However, in practical applications, there may be cases where some modalities are missing [2] or modality data are scarce [3], thus, incomplete modalities bring new challenges to multimodality learning tasks. Incomplete multimodal learning methods aim to handle any available modality during model inference. The segmentation of brain tumors in incomplete multimodality MRI involves the problem of segmenting brain tumor regions from some modality-missing brain tumor images. Segmenting tumor regions from incomplete brain tumor data is more clinically meaningful compared to brain tumor segmentation tasks with complete modalities. However, in some cases, certain modalities may not be available due to technical issues (e.g., image corruption, visual artifacts, imaging protocols), patient allergies to contrast agents, or financial constraints. Segmentation of brain tumors in MRI with missing modalities poses a significant challenge due to the incomplete information available. Researchers and clinicians have developed specialized algorithms and techniques to address this challenge.

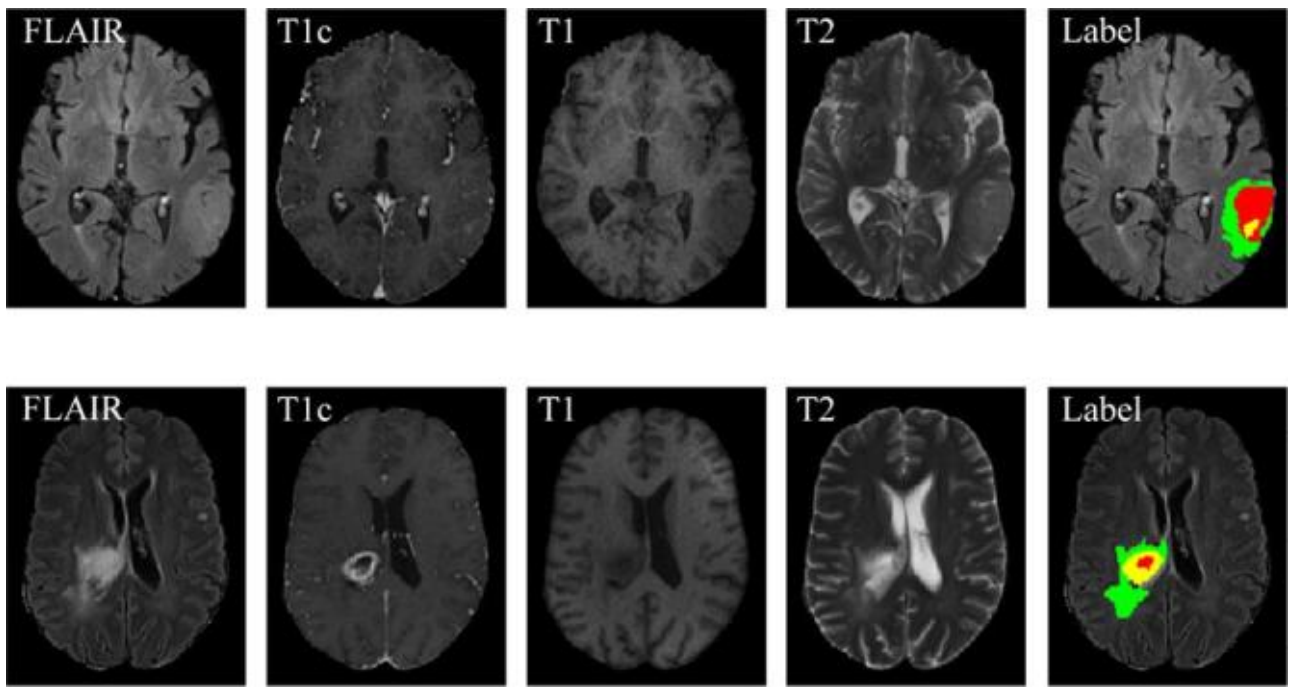

Fig. 1. Examples of four-modality MRI images. From left to right, there are FLAIR images, T1ce images (T1ce, using contrast agent in T1 images), T1-weighted images (T1), T2-weighted images (T2), and images with ground truth labels that contain four sections: black area (background and healthy tissue), green area (enhancing tumor), red area (gangrene and nonenhancing tumor), and yellow area (peritumoral edema).

Segmentation methods that can adapt to incomplete modalities provide flexibility in clinical practice, ensuring consistent and reliable results across different scenarios. Methods examined in the previous studies include the classical synthesis techniques alongside the more recent strategies that utilize deep learning. The newer approaches include common latent space models, knowledge distillation networks, mutual information maximization, and generative adversarial networks [4], [5]. The issue of missing modalities presents a significant challenge that needs to be addressed when conducting brain tumor segmentation tasks in multimodal MRI. The approaches of brain tumor segmentation in MRI with missing modalities previously involved leveraging the available modalities effectively, incorporating classical convolutional neural network (CNN) approaches, such as 3-D U-Net [6] series, and utilizing Transformer [7] and CNN-Transformer methodologies [8] to compensate for missing imaging modalities. In recent years, Mamba [9], [10], derived from state space models (SSMs), is becoming increasingly popular in computer vision [11] due to its ability to handle long-range dependencies effectively while offering greater computational efficiency compared to Transformers.

In this article, we propose a multimodal feature distillation (MFD) with a Mamba-Transformer hybrid network for incomplete multimodal brain tumor segmentation (MMTSeg). First, we design an MFD module that uses a multimodal encoder network to distill complementary multimodal knowledge into the unimodal encoders to extract modality-specific features that are robust to missing modalities. Second, we develop a novel unimodal feature enhancement (UFE) module that incorporates an SSM block into the Transformer as an adapter of the network to introduce long-range dependencies without affecting the Transformer's global modeling capability. Third, we develop a cross-modal fusion (CMF) module that enables the fusion of feature representations from different modalities, even when some modalities are missing. It also employs a transformer-based architecture to establish inter-modal relationships and incorporates a Mamba block to learn the long-range dependencies of features shared among modalities. Besides, the generalized surface loss (GSL) function [12], which pushes the model to reduce the surface distance between predicted and true contours, is employed as the segmentation loss of the proposed MMTSeg to improve boundary-aware feature propagation in representation learning for brain tumor segmentation with incomplete MRI modalities, where high boundary accuracy is important.

The main contributions of this work are summarized as follows.

1) To our best knowledge, MMTSeg is the first Mamba-Tranformer model for incomplete multimodal representation of brain tumor segmentation. By integrating Mamba blocks into Transformers, the unimodal feature extraction process can dynamically capture spatial information and model long-range dependencies on global contextual understanding with linear complexity.
2) Our proposed MMTSeg also has a novel hybrid Mamba-Tranformer for cross-modal learning of brain tumor segmentation, which is designed to capture both spatial and global dependencies for complementary information within and across different modalities and compensate for the absence of certain modalities. Incorporation of Mamba blocks in the proposed CMF across different modal features is adept at linear scaling in sequence length for each modality of the brain tumor MRI, which compensates for Transformers' computational inefficiency on long sequences.
3) We first introduce a boundary-wise loss function for missing modalities segmentation, where boundary information often matters most due to low contrast in unimodality. We conduct extensive experiments on the BraTS 2018 and BraTS 2020 brain tumor datasets and show that our method achieves better performance than some state-of-the-art methods in situations of missing modalities.

## II. Related Work

### A. Feature Distillation in Knowledge Distillation

Knowledge distillation is to get the student network trained by minimizing the distillation loss, which realizes knowledge transfer between teacher and student models that are isomorphic or nonisomorphic networks. In contrast to transferring label knowledge from the teacher network [13], feature distillation, as another approach of knowledge distillation, utilizes intermediate representations learned by the teacher network as hints to improve the final performance of the student network [14]. Intermediate representation distillation is more effective than label knowledge distillation, as it enhances the representation ability and information volume of transferred knowledge. The classic methods of isomorphic feature distillation, which do not need size matching of feature maps, include FitNets [14], attention transfer [15], probabilistic knowledge transfer [16], factor transfer [17], overhaul [18], and self-distillation [19] among others. The self-distillation process enables the neural network to distill knowledge within itself.

With the development of distillation techniques, some advanced feature distillation methods have been applied to medical images. Contrastive representation distillation [20] combines a contrastive objective measuring the mutual information between the representations learned by a teacher and a student network for feature distillation. Xing et al. [21] employed pathology-genomic knowledge distilled from discrepancy-induced contrastive distillation between the teacher's and student's features to improve the accuracy of glioma grading.

### B. Brain Tumor Segmentation With Incomplete Modalities

Since 2016, incomplete MMTSeg methods have focused on end-to-end encoder-decoder network architecture approaches that handle any subset of available modalities, rather than training multiple networks. The hetero-modal image segmentation (HeMIS) network [22] learned the embedding of multimodal information by computing the mean and variance of the features of any available modality. Shen and Gao [23] treated different missing modalities as a specific domain, employed two segmentation networks to segment the complete multimodal and the missing modal images respectively, and used confrontation learning in the segmentation device to project these features into the same common space. However, it is challenging to align the different distributions when a large number of modalities are missing. Zhang et al. [24] proposed an adaptive feature integration method to learn to segment multimodalities using an unimodal model, but this method can only handle cases where pairs or multiple modalities are available. mmFormer [25] employed Transformers to build intra-modal and intermodal relationships to align global feature representations between different modalities. Token Merging transFormer (TMFormer) [26] interprets information extraction from individual modalities as sequences of tokens with varying and adaptable lengths and then fuses them with an adaptive token merging strategy. Zhang et al. [27] introduced the learnable sorting SSM with CNN to maximize the utilization of available modalities for feature enhancement. All existing designs of feature representations utilize CNN or Transformer for multimodal learning. Our approach differs from previous approaches in that we utilize Mamba-Transformer to not only model the relations within modalities but also consider the interactions between modalities.

In addition to feature completion, recent approaches have explored the use of multimodal feature reconstruction [28], [29], [30], latent space representation learning [31], [32], [33], [34], [35], [36], [37], [38], [39], [40], [41], [42], [43], and enhanced feature fusion [44], [45], [46], [47], [48], [49] for robust MMTSeg. Dorent et al. [28] proposed a hetero-modal variational encoder-decoder (U-HVED) network for joint modality completion and segmentation tasks via regularizing modality reconstruction. $M^3$AE [29] masks both random modalities and patches of the remaining modalities for a reconstruction task to compensate for missing MRI sequences by synthesizing them in the original volume space. $D^2$-net [35] consists of a modality disentanglement stage with the spatial-frequency jointly modality contrastive learning and a tumor-region disentanglement stage with the affinity-guided dense tumor-region knowledge distillation. RFNet [44] is a Region-aware Fusion Network taking the relations between modalities and regions into account for incomplete MMTSeg. Modality-masked fusion Transformers with learnable fusion tokens in $M^2$FTrans [46] minimize the negative influence of missing modalities. IM-Fuse [49] integrates Mamba-based fusion blocks in the skip connections between encoders and decoders to enhance overall segmentation performance for brain tumor segmentation with incomplete modalities. The existing feature completion network employs the feature-connected entanglement mechanism to distinguish between the common and individual characteristics of different modalities with complex interactions, while the CMF network utilizes the gating fusion mechanism to integrate multiple entangled features. It primarily addresses the issues of inconsistency and noise in multimodal images; however, when only a few modalities are available, the model's performance degrades significantly because using only one or two modalities may not be sufficient to provide reliable features for the missing modality.

Incomplete MMTSeg via knowledge distillation is to synthesize missing modalities with a segmentation network with complete modality information [35], [50], [51], [52], [53], [54], [55], [56], [57], [58], [59], [60], [61], [62], [63], [64], [65]. MMCFormer [58] utilizes global contextual agreement modules in each scale of the encoders for knowledge distillation to transfer modality-specific representations. These methods have achieved a certain level of performance, but they need complicated methods to train specific models for each subset of missing modalities, which renders them less suitable for clinical practice. To our knowledge, feature distillation in brain tumor segmentation has not been explored in the existing literature.

## III. Methods

Fig. 2 illustrates an overview of the proposed MMTSeg network for brain tumor segmentation with missing modalities. We construct our networks based on a 3-D U-Net, which consists of encoder and decoder stages for multimodal learning. We first design modality-specific and complete multimodal CNN encoders and decoders to extract discriminative features within each modality and across all modalities, respectively. A set of convolutional encoders and decoders is also designed to perform progressive upsampling of the latent feature space, producing robust segmentation. Additionally, the encoders are trained to segment each modal MR image separately, thereby facilitating the learning of receptive field features that contain local contextual information. This design is commonly employed by the state-of-the-art models of incomplete brain tumor segmentation.

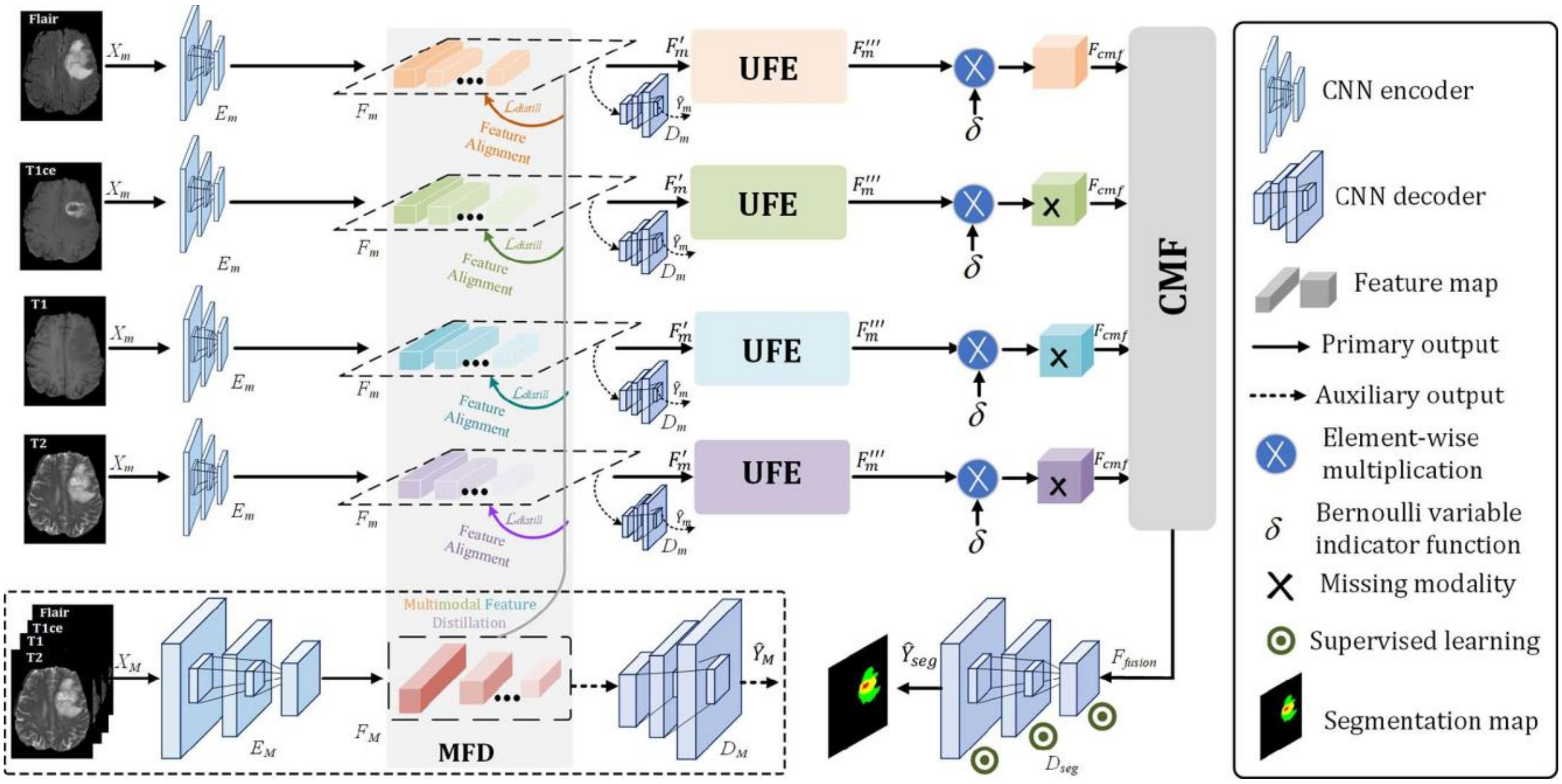

Fig. 2. Overview of the proposed model. The MFD module adopts a multimodal encoder network to distill complementary multimodal knowledge into the unimodal encoders. The UFE module extracts intra-modal feature maps. The CMF module is then used to aggregate features across modalities. $D_{seg}$ produces the final segmentation result $\hat{Y}_{seg}$. The auxiliary outputs $\hat{Y}_m$ of $D_m$ and $\hat{Y}_M$ of $D_M$ are used in loss functions along with the segmentation map of the primary output. The description of symbols in the figure is given in the legend box.

The abovementioned convolutional encoder and decoder are employed in multimodal and unimodal learning for feature self-distillation. The MFD module is composed of student and self-teacher networks, and the student network mimics the feature knowledge distilled by the self-teacher network during the training process. The feature distillation process transfers refined feature knowledge from complete multimodality in the self-teacher network to each unimodal feature representation in the student network. For unimodal learning, feature maps produced by the CNN encoder are refined by the MFD module through self-distillation, where the unimodal training is guided by the complementary knowledge of the complete modal MRI. In the rest of the proposed network, the feature enhancement and fusion modules use each unimodal representation that learned features from well-labeled MRI to handle incomplete modalities.

We introduce two novel modules of feature enhancement and fusion in the encoder-decoder architecture to enhance the model's robustness to incomplete modalities.

1) A set of intra-modal UFEs—computational blocks that combine a Transformer and an Mamba block, for joint learning of dynamically global dependency within each modality.
2) A CMF for multimodal feature fusion to build long-range spatial relationships across different modalities.

The proposed UFE and CMF allow for further enhancement of unimodal and cross-modal features in the presence of missing modalities. By leveraging the complementary strengths of Transformer and Mamba, they respectively enable both modality-specific representation learning within individual modalities, and effective modeling of long-range cross-modal dependencies, resulting in modality-invariant features. Specifically, the UFE module within the MMTSeg framework semantically models the relationship between global and local information for each imaging modality. This module aims to improve the quality of unimodal features and ensure robust feature extraction by each unimodal encoder, even if other modalities are absent during inference. The CMF module integrates features from various MRI modalities, including FLAIR, T1, T1ce, and T2, and manages cases where one or more modalities are absent. The combination of UFE and CMF together extracts high-quality, task-specific features from individual modalities and compensates for missing data.

As an important component of the proposed network, the GSL function [12] yields segmentation outputs for both complete and incomplete modalities representations with more precise boundaries, particularly helpful when tumor margins are complex or subtle. In brain tumor segmentation with missing modalities, models must often infer shape and structural features from less information than a complete set of magnetic resonance sequences provides. With incomplete modalities (e.g., missing FLAIR or T1 for MRI), the model loses contrast and structural cues that help define tumor boundaries. The boundary emphasis in the segmentation loss encourages the model to learn more precise contours, rather than just region overlap, thereby alleviating the negative impact of a lack of contrast and structural cues in a missing-modality setting. Traditional pixel-wise (e.g., cross-entropy) and region-wise (e.g., Dice) losses often fail at image boundaries in medical image segmentation, where misclassification is most common. In contrast, boundary-wise loss has been validated as capable of improving performance and boosting convergence speed for medical image segmentation [66].

### A. Encoder-Decoder Architecture

Let $M = \{FLAIR, T1ce, T1, T2\}$ be the complete set of modalities. We denote the complete multimodal 3-D MRI images by $X_M \in \mathbb{R}^{C\times H\times W\times D}$ where $H \times W$ is the size of spatial resolution, $C$ is the number of modalities, and $D$ is the number of slices, and the data for each modality by $X_m \in \mathbb{R}^{1\times H\times W\times D}, m \in M$. Given the complete input $X_M$, we first design a convolutional encoder $E_M$ to extract compact multimodal features $F_M = E_M(X_M)$. The multimodal encoder $E_M$ consists of five feature extraction layers, each composed of 3-D convolution with $3 \times 3 \times 3$ convolution kernel, instance normalization, and LeakyReLU. Simultaneously, given the unimodal inputs $X_m$, a set of modality encoders $E_m, m \in M$ are adopted to extract modality-specific features $F_m = E_m(X_m)$ for each modality $m$, separately. The unimodal encoders $E_m$ have the same network architecture as $E_M$. Besides, the convolutional decoders have the symmetric path of the convolutional encoders, similar to a U-shape structure.

Then, the MFD module distills complementary multimodal knowledge to improve the training of unimodal encoders by aligning the feature map $F_M$ of the multimodal encoder and $F_m$ of the unimodal encoders to produce $F'_m$ for each modality. The resulting output $F'_m$ is mapped to the same space as $F_M$. To constrain the feature extraction process of the encoders, a multimodal decoder $D_M$ and a set of modality-specific decoders $D_m$, $m \in M$ which correspondingly have five layers, takes $F_M$ and $F'_m$ respectively as inputs to obtain the predicted segmentation maps $\hat{Y}_M$ and $\hat{Y}_m$. The distilled unimodal features $F'_m$ are then fed into the UFE to obtain feature maps $F''_m$, which encode both the local and global contextual information within each modality. The CMF is then used to aggregate features $F''_m$ of all available sets of modalities to obtain fused representations $F_{fusion}$ that capture cross-modal long-range dependency even when some modalities are missing. Finally, a decoder $D_{seg}$ is used to obtain the final segmentation result $\hat{Y}_{seg}$ from the fused features $F_{fusion}$ of incomplete multimodalities.

### B. MFD Module

The proposed MMTSeg melds complementary knowledge via self-distillation for multimodal learning of brain tumor segmentation, which employs logit distillation, likewise M$^3$AE. A notable difference between the two is that M$^3$AE distills the shared semantics between heterogeneous missing-modal situations, whereas MMTSeg distills feature knowledge from complete modality to enhance modality-specific features, guiding fearing learning for unimodality. The MFD for knowledge distillation enables the transfer of feature-level knowledge from the multimodal network (teacher) to unimodal networks (students), providing complementary information to the latter when dealing with missing modalities. It is an end-to-end network that allows for the transfer of multimodal information to unimodal feature extractors or encoders. It can extract representative features for each imaging modality and learn the relationships between different modality representations to increase robustness against cases where modalities are missing.

Integration of features learned from multimodal MRIs can provide complementary information for anatomies and lesion areas for accurate brain tumor segmentation. However, features extracted from individual modality encoders are vulnerable to situations where modality data is missing, which can degrade the subsequent fused representation and, consequently, the segmentation performance. To overcome this issue, we propose an MFD at the encoder stage to transfer complementary knowledge from different modalities in the multimodal teacher encoder, thereby assisting the learning of unimodal student encoders in producing features that are robust against missing modalities.

Let $F^l_M$ and $F^l_m$ be the $l$th layered feature maps of the multimodal encoder $D_M$ (teacher) and the unimodal encoders $D_m$ (students), respectively. To achieve the multimodal knowledge transfer, the feature maps of the student networks are aligned to the teacher network by minimizing a feature-level distillation loss (called MFD loss)

$$\mathcal{L}_{\mathrm{MFD}} = \sum_{m\in M} \sum_{l} \| F^l_M - F^l_m \|_1 \tag{1}$$

where $\|\cdot\|$ denotes the $\ell 1$ norm. The unimodal student encoders can learn multimodal knowledge from the layers of the multimodal teacher network by reducing the discrepancy in the intermediate feature maps, as in (1). Due to the same architectures of the multimodal and unimodal encoders, the MFD is considered an isomorphic feature distillation. Besides, only the high-level semantic features from the middle layers of the unimodal encoders (i.e., $l = 3, 4,$ and 5) are involved in the MFD according to (1). This is to maintain the local invariance property of low-level features that are specific to each modality.

### C. UFE Module

For each modality-specific encoder, we further design a UFE module that combines a Transformer and a Mamba block to extract both global and spatial contextual information within a modality. The use of Transformers in UFE can overcome the limited spatial representation of the generic Transformer encoder and establish a dynamic relationship within a specific modality. Considering that the Transformer has no complementary information specific to vision tasks, we further incorporate the Mamba block at the beginning of the network to offer a better spatial bias into the Transformer. Concretely, the UFE is a hybrid of Transformer and Mamba block arranged in a serial connection structure as shown in Fig. 3(a). In contrast to ViT, which flattens the image into a 1-D token sequence, here we retain the 3-D structure of the input features after the Mamba operation. Given the distilled modality-specific features $F'_m$ as input, UFE produces feature maps $F'''_m$ for each modality $m$ as follows:

$$F''_m = F'_m + MHSA[LN(F'_m)] + Mamba[LN(F'_m)] \quad (2)$$

$$F'''_m = F''_m + MLP[LN(F''_m)] \quad (3)$$

where $MHSA(\cdot)$ is the MultiHead Self-Attention, $MLP(\cdot)$ is MultiLayer Perceptron, and $LN(\cdot)$ is Layer Normalization.

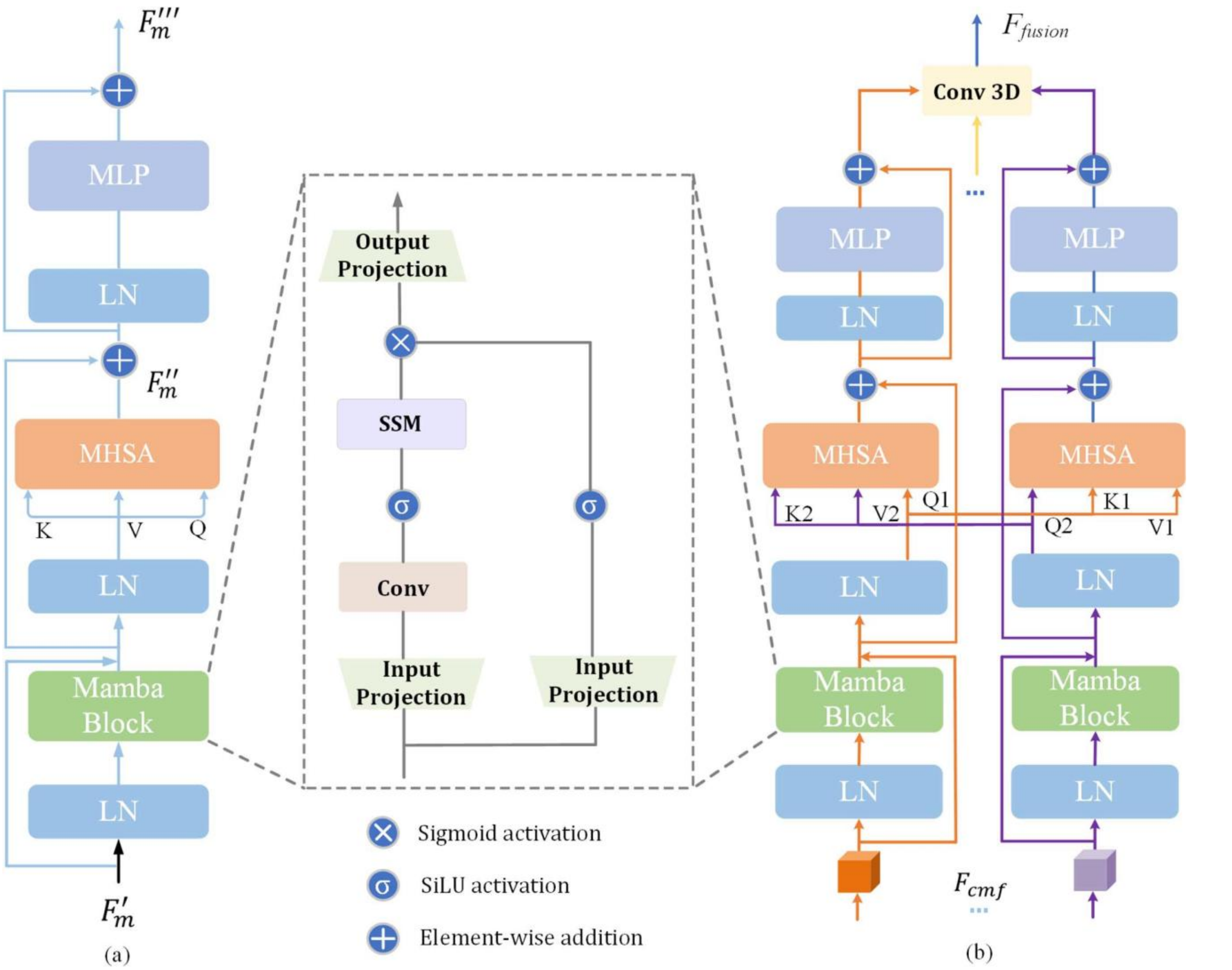

Fig. 3. Structure of: (a) UFE and (b) CMF. The structure of the Mamba block contains the projection layer, the SSM, the convolutional layer, and sigmoid linear unit (SiLU) activation functions, which are in the middle of the figure shared with UFE and CMF.

In contrast to prior Transformer-based segmentation networks, the proposed Mamba-Transformer hybrid network enables unimodal feature extraction and CMF, establishing long-range dependencies among different modalities via the Transformer, while balancing the spatial bias from the Mamba, which further improves segmentation.

## D. CMF Module

We design a CMF module to learn long-range correlations between various modality-specific features, which provide complementary information to improve segmentation. The extracted fused cross-modal features via CMF can represent modality-robust information in the latent space, thereby improving segmentation in settings with missing modalities. The CMF consists of two parallel branches, as shown in Fig. 3(b). In contrast to the intra-modal transformers in UFEs, the first branch is an inter-modal Transformer to learn modality-correlated feature representations. The second branch uses the Mamba block to improve the spatial bias of the inter-modal Transformer in the latent space. Specifically, CMF takes as input varying subsets $\{F'''_m \otimes \delta_m\}$ of available modality-specific features $F'''_m \in \mathbb{R}^{C\times N}$ from UFEs, defined as follows:

$$F_{CMF} = \{F'''_{FLAIR} \otimes \delta_{FLAIR}, F'''_{T1ce} \otimes \delta_{T1ce}, F'''_{T1} \otimes \delta_{T1}, F'''_{T2} \otimes \delta_{T2}\} \quad (4)$$

where $\otimes$ denotes element-wise multiplication and $\delta_m \in \{0, 1\}$ is a binarily-controlled parameter for modality dropout, which represents the presence/absence of a particular modality. Modality features are randomly dropped by setting the corresponding $\delta_m = 0$ to stimulate missing modality. This random modality dropout is implemented during model training to improve the model's robustness against missing modalities.

The $F_{CMF}$ is then processed by the inter-modal Transformer to build long-range dependency across modalities. For example, the resulting cross-modal feature maps between two modalities $\alpha$ and $\beta$ are defined as follows:

$$F_{\alpha\leftarrow\beta} = MHSA(Q_\alpha \cdot K_\beta \cdot V_\beta) = softmax\left(\frac{Q_\alpha \cdot K_\beta^T}{\sqrt{d_k}}\right) V_\beta \tag{5}$$

where $d_k$ represents queries and keys of dimension, and $Q_\alpha$, $K_\beta$ and $V_\beta$ are formulated as follows:

$$Q_\alpha = LN(F_\alpha^{'''}) W_\alpha^Q \tag{6}$$

$$K_\beta = LN(F_\beta^{'''}) W_\beta^K \tag{7}$$

$$V_\beta = LN(F_\beta^{'''}) W_\beta^V. \tag{8}$$

The four modality features in $F_{CMF}$ are integrated by the inter-modal Transformer to get modality-correlated features $F_{trans}$. In parallel, the Mamba block processes the concatenated features of all modalities $Concat(F_{CMF})$ to get $F_{conv} = Mamba[Concat(F_{CMF})]$. The final output of CMF is a fused representation defined by

$$F_{fusion} = F_{trans} \oplus F_{trans} \tag{9}$$

where $\oplus$ denotes element-wise addition. Finally, the fused feature $F_{fusion}$ with 3-D convolutional feature processing is fed to the convolutional segmentation decoder $D_{seg}$ through stepwise upsampling operations to obtain the final segmentation map $\bar{Y}_{seg}$.

### E. Loss Function

We employ a boundary-based loss function for all decoders in the proposed MMTSeg to potentially improve accuracy, as contextual boundary information is crucial for predicting uncertainty estimation in medical image segmentation [67], [68]. On the contrary, Dice loss is only region-based and has an intrinsic bias towards specific extremely class-imbalanced conditions [69] that exists in brain MRI segmentation [70].

To align the prediction with ground-truth segmentation, the unimodal decoders $D_m$ are optimized based on the GSL function [12] as follows:

$$\mathcal{L}_{um} = \sum_{m\in M} \mathcal{L}_{GSL}(\bar{Y}_m, Y_m) \tag{10}$$

where $\bar{Y}_m$ and $Y_m$ are, respectively, the predicted and ground-truth segmentation labels for each modality $m$.

The multimodal decoder $D_M$ is optimized by

$$\mathcal{L}_{mm} = \mathcal{L}_{GSL}(\bar{Y}_M, Y_M) \tag{11}$$

where $\bar{Y}_M$ and $Y_M$ are the predicted and ground truth of complete multimodal images, respectively.

The loss function of the cross-modal decoder $D_{seg}$ is formulated as follows:

$$\mathcal{L}_{seg} = \mathcal{L}_{GSL}(\bar{Y}_{seg}, Y_M). \tag{12}$$

In addition, we adopt a deep supervision strategy to regularize the output features of the first to the fourth layer of the decoder $D_{seg}$. The layer-wise prediction loss is formulated as follows:

$$\mathcal{L}_{layer} = \sum_{l=1}^{4} \mathcal{L}_{GSL}(\bar{Y}_l, Y_l) \tag{13}$$

where $l$ denotes the index of layer in the decoder $D_{seg}$.

The overall loss function is defined as follows:

$$\mathcal{L}_{total} = \mathcal{L}_{um} + \mathcal{L}_{mm} + \mathcal{L}_{seg} + \mathcal{L}_{layer}. \tag{14}$$

## IV. Experiments and Results

We evaluate the effectiveness of our MMTSeg framework for brain tumor segmentation from MRI with missing modalities.

### A. Datasets

We conducted extensive experiments on the 2018 Brain Tumor Segmentation Challenge BraTS 2018 and BraTS 2020 [71] datasets. The BraTS 2018 dataset comprises scan images of 285 patients, while the BraTS 2020 dataset contains scan images of 369 patients. Each sample contains MRI scan images of four different modalities: FLAIR, T1ce, T1, and T2. The ground-truth labels are divided into four categories, including healthy tissue, edema area, necrosis area, and enhancing tumor area. We evaluate segmentation performance on enhancing tumor region (ET), tumor core region (TC), and whole tumor region (WT). WT is composed of an enhancing area, edema area, and necrosis area; TC is composed of an enhancing area and necrosis area; and ET is composed of an enhancing area. The enhancing tumor is a critical sub-region for understanding tumor growth and for planning treatments, as it signifies active tumor tissue, which is a significant challenge in automated brain tumor segmentation. The acquired dataset has been preprocessed by registration, skull removal, and resampling to 1 $mm^3$ resolution.

### B. Evaluation Metrics

While the Dice score is widely used for evaluating brain tumor segmentation, it only focuses on overall region overlap, but doesn't account for the precise boundary details of brain tumors in MRI, such as the location or spatial arrangement of the segmented tumor. This means the Dice score cannot differentiate between biases in segmenting different parts of a brain tumor, which can be crucial in brain MRI. Besides, the Dice score doesn't directly incorporate true negatives (correctly identified negative pixels) into its calculation, leading to disproportionate penalization of small tumors. A small-sized brain tumor with a few false positives can result in a very low Dice score, even if the majority of other pixels were predicted correctly, due to scale sensitivity and bias against small objects in the Dice score.

However, Hausdorff distance (HD) measures how closely the predicted boundary aligns with the true anatomical boundary, which is critical for precise tumor delineation, especially in surgical or radiation planning, including brain tumor segmentation.

TABLE I
COMPARISON OF BRAIN TUMOR SEGMENTATION PERFORMANCE BETWEEN MMTSEG AND OTHER STATE-OF-THE-ART CNN/TRANSFORMER-BASED MODELS ON THE BRATS 2018 DATASET BY BAHD (MM)

| | M | | | | | | | | | | | | | | | | Avg |
|---|---|---|---|---|---|---|---|---|---|---|---|---|---|---|---|---|---|
| | FLAIR | ● | ● | ● | ✚ | ● | ● | ✚ | ● | ✚ | ✚ | ✚ | ✚ | ✚ | ● | ✚ | |
| | T1 | ● | ● | ✚ | ● | ● | ✚ | ✚ | ✚ | ● | ● | ✚ | ✚ | ● | ✚ | ✚ | |
| | T1ce | ● | ✚ | ● | ● | ✚ | ✚ | ● | ● | ● | ✚ | ✚ | ● | ✚ | ✚ | ✚ | |
| | T2 | ✚ | ● | ● | ● | ✚ | ● | ● | ✚ | ✚ | ● | ● | ✚ | ✚ | ✚ | ✚ | |
| WT | HeMIS [22] | 23.70 | 57.13 | 85.51 | 16.07 | 15.89 | 42.10 | 14.72 | 20.01 | 14.10 | 12.35 | 12.33 | 13.61 | 13.20 | 13.02 | 12.13 | 24.39 |
| | U-HVED [28] | 31.71 | 30.92 | 32.29 | 33.26 | 24.58 | 21.09 | 22.12 | 24.23 | 26.35 | 21.93 | 22.47 | 18.53 | 13.64 | 12.38 | 11.17 | 23.18 |
| | RobustMSeg [31] | 12.80 | 18.20 | 20.33 | 10.67 | 12.40 | 9.73 | 10.34 | 10.56 | 10.45 | 11.18 | 9.76 | 9.52 | 9.91 | 10.01 | 9.36 | 11.68 |
| | $D^2$-Net [35] | 16.50 | 13.38 | 17.17 | 12.29 | 16.53 | 11.49 | 10.95 | 11.93 | 10.86 | 10.56 | 11.73 | 9.69 | 9.66 | 9.39 | 9.21 | 12.09 |
| | RFNet [44] | 6.79 | 8.63 | 8.45 | 5.30 | 4.28 | 7.51 | 7.16 | 6.91 | 5.02 | 3.56 | 3.74 | 3.25 | 3.47 | 4.16 | 3.35 | 5.43 |
| | mmFormer [25] | 5.76 | 6.61 | 7.73 | 7.42 | 5.31 | 5.46 | 5.34 | 5.25 | 5.69 | 5.42 | 5.98 | 5.69 | 5.28 | 5.49 | 5.03 | 5.83 |
| | $M^3$AE [29] | 5.85 | 7.16 | 7.87 | 4.87 | 5.03 | 5.96 | 5.11 | 5.14 | 5.01 | 5.09 | 3.68 | 3.92 | 3.37 | 4.18 | 3.66 | 5.06 |
| | **MMTSeg (Ours)** | **4.45** | **7.15** | **8.42** | **2.77** | **3.93** | **6.45** | **3.10** | **4.89** | **2.67** | **3.11** | **3.35** | **2.85** | **2.94** | **3.34** | **2.79** | **4.14** |
| TC | HeMIS | 31.63 | 27.31 | 31.47 | 35.31 | 19.64 | 30.58 | 28.45 | 25.07 | 29.21 | 19.18 | 18.31 | 24.12 | 25.98 | 18.38 | 18.12 | 25.51 |
| | U-HVED | 29.14 | 32.57 | 34.39 | 36.34 | 26.09 | 19.18 | 31.63 | 29.55 | 29.49 | 18.89 | 17.74 | 18.39 | 16.70 | 16.02 | 15.18 | 24.75 |
| | RobustMSeg | 10.29 | 12.00 | 19.33 | 10.60 | 10.12 | 12.58 | 8.24 | 10.57 | 9.45 | 11.18 | 8.76 | 8.52 | 8.55 | 10.15 | 7.75 | 10.02 |
| | $D^2$-Net | 16.54 | 18.40 | 18.37 | 17.25 | 15.33 | 12.02 | 14.61 | 14.45 | 12.82 | 12.00 | 11.96 | 11.34 | 10.38 | 11.64 | 10.11 | 13.81 |
| | RFNet | 7.63 | 5.31 | **7.47** | 7.31 | 6.64 | 7.58 | 6.45 | 6.07 | 7.21 | 6.18 | 5.31 | 6.12 | 5.98 | 6.38 | 5.12 | 6.45 |
| | mmFormer | **7.19** | 6.04 | 7.83 | 7.81 | 5.21 | 7.07 | 6.74 | 7.02 | 6.45 | 6.07 | 7.10 | 6.18 | 5.31 | 6.88 | 5.14 | 6.54 |
| | $M^3$AE | 7.66 | **5.29** | 7.49 | 7.41 | 6.55 | 6.62 | 7.08 | 7.15 | 7.07 | 5.83 | 5.91 | 6.30 | 5.34 | 5.26 | 5.10 | 6.40 |
| | **MMTSeg (Ours)** | 9.06 | 7.48 | 14.24 | **6.68** | **4.80** | **6.66** | **6.97** | **7.00** | **5.65** | **3.93** | **4.08** | **6.03** | **3.53** | **3.72** | **3.86** | **6.24** |
| ET | HeMIS | 18.94 | 13.31 | 17.53 | 15.73 | 15.34 | 14.36 | 14.83 | 15.17 | 11.12 | 18.52 | 14.38 | 11.84 | 11.41 | 11.66 | 11.26 | 14.36 |
| | U-HVED | 15.85 | 13.49 | 17.82 | 16.60 | 12.34 | 12.82 | 13.02 | 12.27 | 13.71 | 12.43 | 13.16 | 11.41 | 10.46 | 10.79 | 10.21 | 13.09 |
| | RobustMSeg | 9.31 | 7.01 | 10.51 | 10.93 | 8.07 | 8.24 | 9.52 | 9.45 | 9.05 | 6.87 | 6.36 | 7.62 | 7.30 | 6.10 | 6.08 | 8.16 |
| | $D^2$-Net | 12.62 | 13.51 | 13.67 | 15.30 | 13.40 | 14.26 | 12.93 | 11.53 | 10.97 | 12.13 | 11.30 | 10.15 | 10.41 | 9.72 | 9.17 | 12.07 |
| | RFNet | 5.32 | 6.67 | 8.84 | 5.31 | 5.46 | 5.95 | 5.96 | 5.80 | 4.75 | 4.34 | 4.99 | 4.18 | 4.39 | 3.64 | 3.49 | 5.27 |
| | mmFormer | **4.91** | 5.40 | 8.47 | 7.33 | 7.00 | 7.14 | 7.48 | **4.56** | 4.51 | 4.07 | 4.43 | 4.67 | 6.70 | 5.15 | 4.39 | 5.75 |
| | $M^3$AE | 5.56 | 6.13 | 7.52 | 5.27 | 5.29 | 5.38 | 5.72 | 5.77 | 5.38 | 4.44 | 4.31 | 4.64 | 4.28 | 4.02 | 3.98 | 5.18 |
| | **MMTSeg (Ours)** | 5.01 | **4.49** | **6.08** | **3.92** | **3.35** | **3.64** | **5.60** | 5.67 | **3.58** | **2.57** | **2.53** | **2.93** | **2.27** | **2.49** | **2.19** | **3.75** |

Note: bAHD (mm) is used to evaluate the performance of the model on the three tumor regions of WT, TC, and ET. M represents modalities, of which value is the smaller ↓ the better. Avg indicates the average value of 14 different missing modalities and 1 complete modality. ● denotes the modality is missing, ✚ denotes the modality exists. The best results are shown in bold.

The formulation of HD between two points X and Y in a 3-D MRI image is presented as follows:

$$HD(Y_c, \hat{Y}_c) = \max\{d(Y_c, \hat{Y}_c), d(\hat{Y}_c, Y_c)\}. \quad (15)$$

Nevertheless, average HD can misrank segmentation results, especially when errors are nonoverlapping or distributed unevenly in the 3-D image segmentation, and do not necessarily correspond to a segmentation of higher quality.

To avoid the disadvantages of the Dice score and the ranking error of the average HD in evaluating medical image segmentation, we employ the balanced Average Hausdorff Distance (bAHD) [72] for a more comprehensive evaluation, which is defined as follows:

$$bAHD(G, S) = \left(\frac{GtoS}{G} + \frac{StoG}{G}\right)/2 \quad (16)$$

where $G$ and $S$ are the number of voxels in the ground truth and the model prediction, respectively. GtoS is the sum of all minimum distances from all points from the ground truth to the model prediction, and StoG is the sum of all minimum distances from all points from the model prediction to the ground truth. The metric of bAHD aims to calculate the maximum distance between corresponding points on the boundary of patches within one or two boundaries in a 3-D image. The smaller the value of the bAHD, the better the segmentation prediction result.

## C. Implementation Details

During the experiment, we performed Z-score normalization on the input MRI image, randomly cropped the spatial resolution of the input image to a size of $128 \times 128 \times 128$, and then performed random rotation, intensity shift, and mirror flip. Adam optimizer was used to optimize the network parameters, $\beta 1$ and $\beta 2$ are 0.9 and 0.999, respectively, and the weight decay is $1e^{-5}$. The entire network was trained for 1000 epochs, and the batch size was set to 1. In addition, we adopted a polynomial learning rate strategy similar to [73], with an initial learning rate of $initial_{lr} = 1e^{-4}$, and a current learning rate of $lr = initial_{lr} \times (1 - num/max_{epoch})^p$ with $p = 0.9$ and $num$ is the epoch number.

## D. Quantitative Results

*1) Evaluation Results on BraTS 2018:* We compare the proposed MMTSeg with state-of-the-art methods on different cases with missing MRI modalities on the BraTS 2018 dataset. These competing methods include HeMIS [22], U-HVED [28], RobustSeg [31], $D^2$-Net [35], RFNet [44], mmFormer [25], and $M^3$AE [29]. As shown in Table I, we use the Dice score as the evaluation index, and the bAHD of the proposed model on WT, TC, and ET are 4.41, 6.24, and 3.75 mm, respectively. Compared with the second and third-best RFNet and mmFormer, our network improves on the WT, TC, and ET regions on average across 15 different modality deletion cases, and the results show that our model obtains the tumor core and

TABLE II
COMPARISON OF BRAIN TUMOR SEGMENTATION PERFORMANCE BETWEEN MMTSEG AND MODELS MODIFIED WITH GSL ON THE BRATS2018 DATASET BY BAHD (MM)

| | Average bAHD Score (mm) ↓ | | |
|---|---|---|---|
| Models | WT | TC | ET |
| mmFormer with GSL | 7.47 | 8.58 | 6.89 |
| M³AE with GSL | 7.66 | 8.23 | 6.57 |
| **MMTSeg (Ours)** | **4.14** | **6.24** | **3.75** |

Note: bAHD (mm) is used to evaluate the performance of the models with GSL function. The best results are shown in bold.

TABLE III
COMPARISON OF BRAIN TUMOR SEGMENTATION PERFORMANCE BETWEEN MMTSEG AND OTHER STATE-OF-THE-ART METHODS ON THE BRATS2018 DATASET BY DICE SCORE (%)

| | Average Dice Score (%) ↑ | | |
|---|---|---|---|
| Models | WT | TC | ET |
| HeMIS | 69.98 | 52.24 | 43.76 |
| U-HVED | 72.21 | 53.46 | 46.24 |
| RobustSeg | 79.98 | 71.28 | 52.01 |
| RFNet | 85.76 | 76.92 | 59.47 |
| mmFormer | 83.59 | 75.23 | 58.08 |
| M³AE | 86.01 | **77.39** | 60.22 |
| **MMTSeg (Ours)** | **87.50** | 72.58 | **64.63** |

Note: Dice score (%) is used to evaluate model performance of the three tumor regions of WT, TC, and ET. The best results are shown in bold.

the overall tumor region. Distinct from the Dice score, all the models obtain a better effect on enhancing tumors (i.e., ET) than whole tumors (i.e., WT) and tumor core (i.e., TC) because the boundary-based metrics, including bAHD, are more sensitive to the hyper-intense and low contrast regions that contain small lesions. Significant improvement, and as can be seen from the bAHD of ET in Table I, our proposed method achieves the best performance among any other models under 15 different modality missing cases. This can be attributed to the capability of our model to adapt to all tumor regions and to establish a dynamic semantic relationship well.

Our method uses the GSL function, which aligns closely with the bAHD metric. For a fair performance comparison in terms of the bAHD metric under the same loss function setting, we compare the proposed MMTSeg with key baselines, i.e., the mmFormer and M³AE modified with the GSL function, where the loss function of the supervised learning segmentation process in mmFormer and M³AE is replaced by GSL. The results are shown in Table II. Nevertheless, the performance of mmFormer and M³AE retrained with GSL drops in terms of increased bAHD, compared to their performance when trained with the original losses shown in Table I. The mmFormer uses a softmax-weighted loss in the convolutional decoder as an auxiliary regularizer to encourage accurate segmentation in the presence of missing modalities. The combination of the softmax-weighted loss and the GSL, which replaces the Dice loss, may lead to lower performance than a well-tuned region-only loss, because the GSL's focus on boundary outliers can induce noise in the Softmax loss. M³AE originally employs Mean Squared Error (MSE) loss for reconstruction and self-distilling consistency and Dice loss for segmentation. The replacement of Dice loss with GSL is likely to cause conflicting optimization targets, because the region-based MSE may dominate the training process, forcing the model to ignore the precise boundary refinement proposed by surface-based GSL.

Given that the Dice score remains the most widely reported metric for the BraTS challenges, we continue to use it as one of the evaluation metrics in our experiments. As shown in Table III, the proposed MMTSeg also shows superior Dice scores for the WT and ET labels, achieving 87.50% and 64.63% on the BraTS 2018 dataset, respectively.

*2) Evaluation Results on BraTS 2020:* To evaluate the robustness of our proposed method, we also conducted experiments on the BraTS 2020 dataset. We compared seven advanced models, including HeMIS, U-HVED, RobustSeg, D²-Net, RFNet, mmFormer, and M³AE. As shown in Table IV, we use bAHD as the evaluation index for the proposed model, with values of 3.65, 5.55, and 3.12 mm for WT, TC, and ET, respectively. Compared to the second- and third-best RFNet and mmFormer, our model outperforms them in the WT, TC, and ET regions, respectively. The results show that our method achieves a significant improvement in the tumor core and enhances tumor regions. Compared with pure CNN or Transformer and their hybrid methods, our proposed Mamba-Transformer hybrid architecture can improve the model's feature representation fusion and semantic understanding, which leads to more accurate segmentation in MRI brain tumor images with missing modalities.

*3) Comparisons for Model Efficiency:* We also compare model efficiency between the proposed MMTSeg and other state-of-the-art models in terms of parameters and floating point operations (FLOPs). Table V shows the model complexity of other Transformer-based models for brain tumor segmentation with complete modalities. In terms of computational efficiency, the proposed MMTSeg outperforms certain CNN-based models, including U-HVED and RobustMSeg. Compared to the lower-computation Transformer-based approach, mmFormer, our proposed MMTSeg has fewer model parameters but more GFLOPs, due to the incorporation of the Mamba block, which involves fewer computations. However, it is reasonable to expect that Mamba blocks bring a substantial improvement in segmentation accuracy, regardless of whether modalities are complete or incomplete.

## *E. Ablation Study*

*1) Effectiveness of Proposed Components:* We conducted a series of ablation studies on our proposed MMSeg. As shown in Table VI, the proposed MFD, UFE, and CMF modules can significantly improve network performance. For example, after removing the complementary knowledge guide module (i.e, MFD), the performance of the three tumor regions of WT, TC, and ET relatively decreased by an average of 23.32%, suggesting that the complementary knowledge guide module can effectively distill multimodal complementary information

TABLE IV
COMPARISON OF BRAIN TUMOR SEGMENTATION PERFORMANCE ON THE BRATS 2020 DATASET BY BAHD (MM)

| M | FLAIR | ● | ● | ● | ✣ | ● | ● | ✣ | ● | ✣ | ✣ | ✣ | ✣ | ✣ | ● | ✣ | Avg |
|---|---|---|---|---|---|---|---|---|---|---|---|---|---|---|---|---|---|
| | T1 | ● | ● | ✣ | ● | ● | ✣ | ✣ | ✣ | ● | ● | ✣ | ✣ | ● | ✣ | ✣ | |
| | T1ce | ● | ✣ | ● | ● | ✣ | ✣ | ● | ● | ● | ✣ | ✣ | ● | ✣ | ✣ | ✣ | |
| | T2 | ✣ | ● | ● | ● | ✣ | ● | ● | ✣ | ✣ | ● | ● | ✣ | ✣ | ✣ | ✣ | |
| WT | HeMIS | 37.13 | 22.05 | 47.71 | 42.72 | 16.45 | 18.49 | 32.59 | 28.99 | 29.35 | 28.65 | 18.07 | 29.36 | 17.97 | 17.53 | 16.83 | 26.93 |
| | U-HVED | 38.92 | 31.40 | 35.22 | 36.12 | 24.81 | 26.66 | 27.75 | 29.09 | 27.62 | 23.58 | 25.42 | 21.01 | 22.80 | 24.70 | 20.05 | 27.67 |
| | RobustMSeg | 13.21 | 20.34 | 21.46 | 12.95 | 18.00 | 11.55 | 11.30 | 11.35 | 12.95 | 12.94 | 9.39 | 9.81 | 9.42 | 10.59 | 9.23 | 12.94 |
| | $D^2$-Net | 15.47 | 15.34 | 17.09 | 18.80 | 14.44 | 15.45 | 14.25 | 13.79 | 12.24 | 11.52 | 11.27 | 10.46 | 10.67 | 9.04 | 9.35 | 13.40 |
| | RFNet | 7.27 | 7.18 | 8.10 | 7.45 | 7.52 | 8.70 | 8.46 | 9.34 | 7.66 | 6.44 | 4.54 | 4.30 | 4.15 | 5.80 | 5.07 | 6.79 |
| | mmFormer | 6.85 | 7.78 | 8.48 | 5.90 | 5.43 | 6.65 | 5.58 | 5.35 | 5.59 | 6.01 | 5.87 | 4.36 | 4.75 | 5.39 | 4.14 | 5.87 |
| | $M^3$AE | 6.79 | **6.07** | **7.13** | 6.10 | 6.49 | 5.99 | 6.08 | 6.16 | 6.27 | 6.84 | 5.15 | 4.68 | 4.54 | 5.32 | 4.11 | 5.85 |
| | **MMTSeg (Ours)** | **3.91** | 6.24 | 7.48 | **2.47** | **3.42** | **5.85** | **2.73** | **4.18** | **2.36** | **2.72** | **2.93** | **2.49** | **2.57** | **2.94** | **2.45** | **3.65** |
| TC | HeMIS | 31.63 | 27.31 | 31.47 | 33.31 | 18.64 | 29.58 | 28.45 | 23.07 | 27.21 | 19.18 | 18.31 | 23.12 | 24.98 | 18.38 | 18.12 | 24.84 |
| | U-HVED | 23.70 | 47.13 | 65.51 | 16.07 | 15.89 | 42.10 | 14.72 | 20.01 | 14.10 | 12.35 | 12.33 | 13.61 | 13.20 | 13.02 | 12.13 | 22.39 |
| | RobustMSeg | 10.93 | 18.74 | 12.90 | 13.26 | 13.95 | 10.02 | 14.69 | 16.41 | 12.81 | 12.17 | 11.37 | 9.24 | 11.17 | 10.02 | 9.70 | 11.83 |
| | $D^2$-Net | 14.12 | 12.15 | 17.04 | 14.49 | 16.90 | 12.18 | 11.32 | 12.84 | 10.24 | 11.89 | 10.74 | 11.39 | 11.70 | 12.02 | 10.18 | 12.61 |
| | RFNet | 7.19 | 6.22 | 7.18 | 7.86 | 6.88 | 7.64 | **6.08** | 6.54 | 7.87 | 5.89 | 6.68 | 5.69 | 5.63 | 5.48 | 5.77 | 6.57 |
| | mmFormer | 7.71 | 7.92 | **7.29** | 7.26 | 6.58 | 6.09 | 6.12 | 6.23 | 6.35 | 6.93 | 6.47 | 5.53 | 5.64 | 5.38 | 5.17 | 6.47 |
| | $M^3$AE | **6.97** | 7.38 | 7.67 | 6.39 | 6.08 | 6.15 | 6.19 | 7.11 | 6.24 | 6.20 | 6.01 | 5.91 | 5.88 | 6.07 | 5.56 | 6.38 |
| | **MMTSeg (Ours)** | 7.80 | **6.32** | 13.27 | **6.10** | **4.20** | **5.85** | 6.18 | **6.14** | **5.16** | **3.47** | **3.60** | **5.31** | **3.16** | **3.25** | **3.40** | **5.55** |
| ET | HeMIS | 12.26 | 18.97 | 30.56 | 13.49 | 13.95 | 17.04 | 20.43 | 13.84 | 17.01 | 15.70 | 17.09 | 10.54 | 11.03 | 17.46 | 11.01 | 15.94 |
| | U-HVED | 14.39 | 15.67 | 13.89 | 13.77 | 12.29 | 19.08 | 18.18 | 16.33 | 15.27 | 14.95 | 13.01 | 12.29 | 11.92 | 12.09 | 11.55 | 14.31 |
| | RobustMSeg | 9.41 | 8.51 | 8.42 | 9.71 | 8.17 | 8.13 | 7.29 | 8.38 | 7.12 | 8.64 | 8.12 | 7.94 | 7.26 | 7.18 | 7.95 | 8.14 |
| | $D^2$-Net | 9.33 | 9.18 | 9.95 | 8.97 | 8.02 | 8.15 | 8.91 | 8.88 | 8.53 | 8.19 | 8.22 | 8.27 | 8.40 | 7.64 | 7.12 | 8.52 |
| | RFNet | 5.13 | 5.05 | 5.71 | 5.72 | 4.45 | 4.49 | 5.59 | 5.35 | 4.89 | 4.65 | 4.07 | 4.36 | 4.97 | 4.83 | 4.53 | 4.88 |
| | mmFormer | 5.51 | 5.02 | 5.59 | 5.81 | 4.01 | 5.22 | 4.81 | 4.96 | 4.95 | 4.51 | 4.24 | 3.66 | 3.81 | 3.41 | 3.08 | 4.57 |
| | $M^3$AE | 5.47 | 4.94 | 5.67 | 5.48 | 3.96 | 4.42 | 4.99 | 4.76 | 4.63 | 4.22 | 4.18 | 3.37 | 3.62 | 3.33 | 3.04 | 4.41 |
| | **MMTSeg (Ours)** | **4.01** | **3.72** | **4.85** | **3.41** | **2.82** | **3.09** | **4.09** | **4.61** | **3.62** | **2.21** | **2.34** | **2.11** | **2.05** | **2.13** | **1.93** | **3.12** |

Note: bAHD (mm) is used to evaluate the performance of the model on the three tumor regions of WT, TC, and ET. M represents modalities, of which value is the smaller ↓ the better. Avg indicates the average value of 14 different missing modalities and 1 complete modality. ● denotes the modality is missing, ✣ denotes the modality exists. The best results are shown in bold.

TABLE V
COMPARISON OF MODEL EFFICIENCY BETWEEN MMTSEG AND OTHER STATE-OF-THE-ART MODELS FOR COMPLETE MODALITY BRAIN TUMOR SEGMENTATION

| Models | Params (M) | FLOPs (G) |
|---|---|---|
| HeMIS | 1.17 | 77.88 |
| U-HVED | 3.79 | 284.37 |
| RobustMSeg | 37.58 | 846.54 |
| RFNet | 25.40 | 204.57 |
| mmFormer | 57.61 | 206.83 |
| $M^3$AE | 27.21 | 196.11 |
| MMTSeg (Ours) | 48.24 | 257.41 |

Note: Parameters (Params) and Floating Point Operations (FLOPs) are used to evaluate model complexity under four modalities. The unit for Params is Mega/million (M), and for FLOPs is Giga/billion (G).

TABLE VI
ABLATION STUDY OF PROPOSED COMPONENTS ON BRATS 2018 DATASET

| Method | Average bAHD (mm) ↓ | | |
|---|---|---|---|
| | WT | TC | ET |
| W/O MFD | 5.82 | 7.54 | 4.53 |
| W/O UFE | 7.20 | 8.64 | 6.26 |
| W/O CMF | 8.54 | 9.27 | 7.43 |
| MMTSeg | 4.14 | 6.24 | 3.75 |

Note: W/O MFD, W/O UFE, and W/O CMF represent removing MFD/UFE and replacing CMF respectively.

to unimodality and improve the representativeness of the unimodal features. After removing the multimodal feature representation module (i.e., UFE), the performance of the three tumor regions of WT, TC, and ET relatively decreased by an average of 36.79%, indicating that the UFE module can well model the global relationship within the modality. After replacing the proposed cross-modal feature fusion module (i.e., CMF) with a 3-D convolutional fusion decoder in the RFNet, mmFormer, and $M^3$AE, the performance of the three tumor regions of WT, TC, and ET decreased by an average of 43.56% relatively. This suggests that the long-range correlation between different modalities should be effectively aligned through the information exchange between modalities to improve segmentation performance.

*2) Effectiveness of Mamba Block:* We also evaluated the effectiveness of the Mamba block. We examine the effect of the proposed UFE by removing the Mamba block from the UFE and CMF. It can be seen from Table VII that, after removing the Mamba block in UFE, the performance of the three tumor regions of WT, TC, and ET decreased by an average of 13.53% relatively. After removing the Mamba block in CMF, the performance of the three tumor regions, ET, TC, and WT, is relatively reduced by an average of 24.80%. This suggests that the introduction of Mamba can compensate for Transformer's insufficient dynamism.

In addition, we assess the effectiveness of the Mamba block by providing ablation results of "W/O MHSA in UFE & CMF"

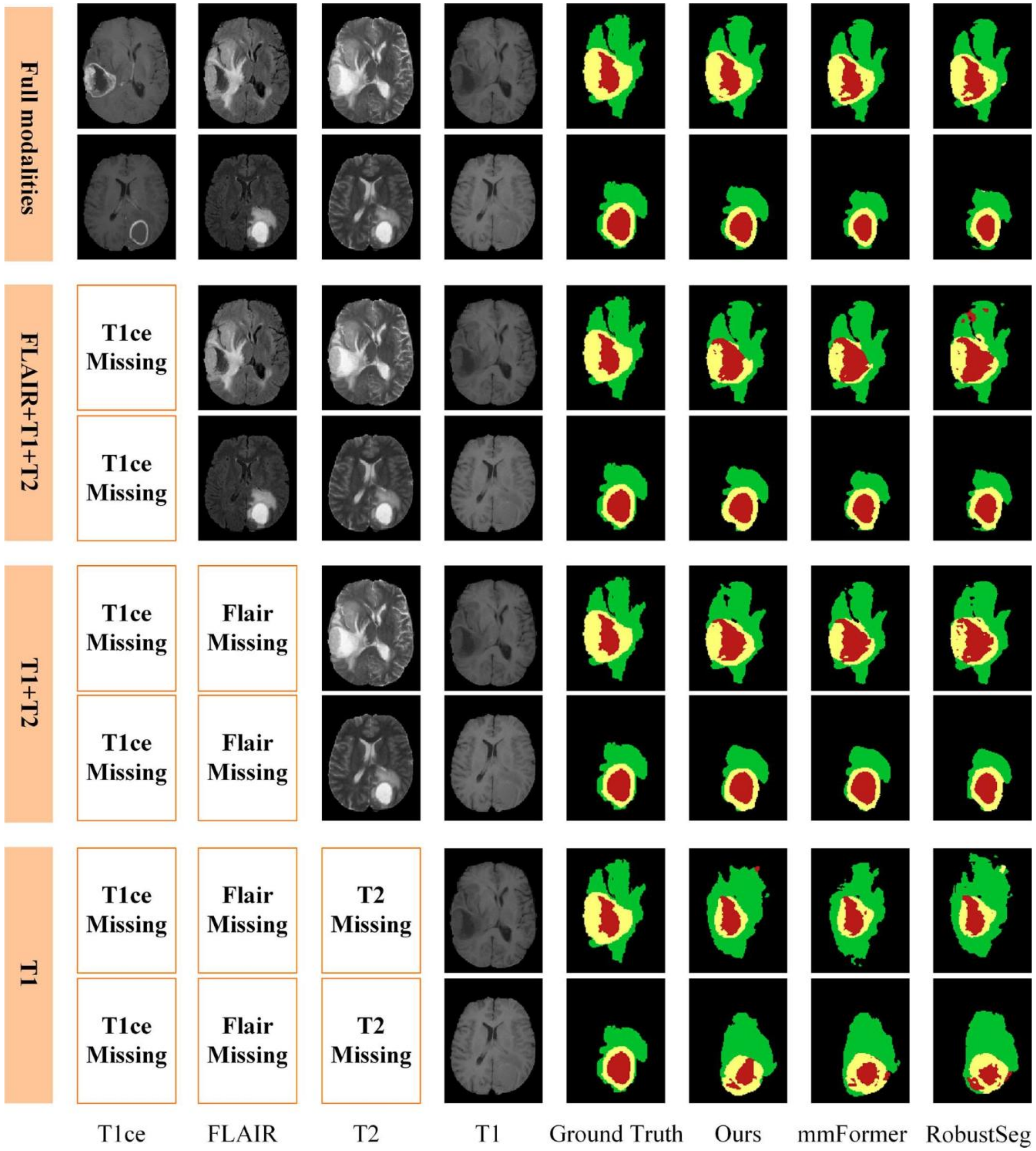

Fig. 4. Visualization of mmFormer, RobustSeg, and our proposed model on the BraTS 2018 dataset. There are more segmented areas in images of our proposed model than in the other two.

(i.e., "pure Mamba") and "W/O Mamba in UFE & CMF" (i.e., "pure Transformer"). Results further confirm that incorporating the Mamba block significantly improves performance, as indicated by substantial drops in bAHD when the Mamba blocks are removed from the UFE and CMF modules. The proposed UFE and CMF, which both interleave Mamba and Transformer blocks, combine the two excellent architectures to leverage their complementary strengths. In the hybrid approaches, Mamba compensates for the scalability issues of Transformers, while Transformer layers can be added to Mamba for improved generalization and local modeling.

*3) Effectiveness of CMF Module:* To validate the effectiveness of our proposed Mamba-Tranformer CMF module, we conduct ablation experiments by replacing CMF with pure CNN

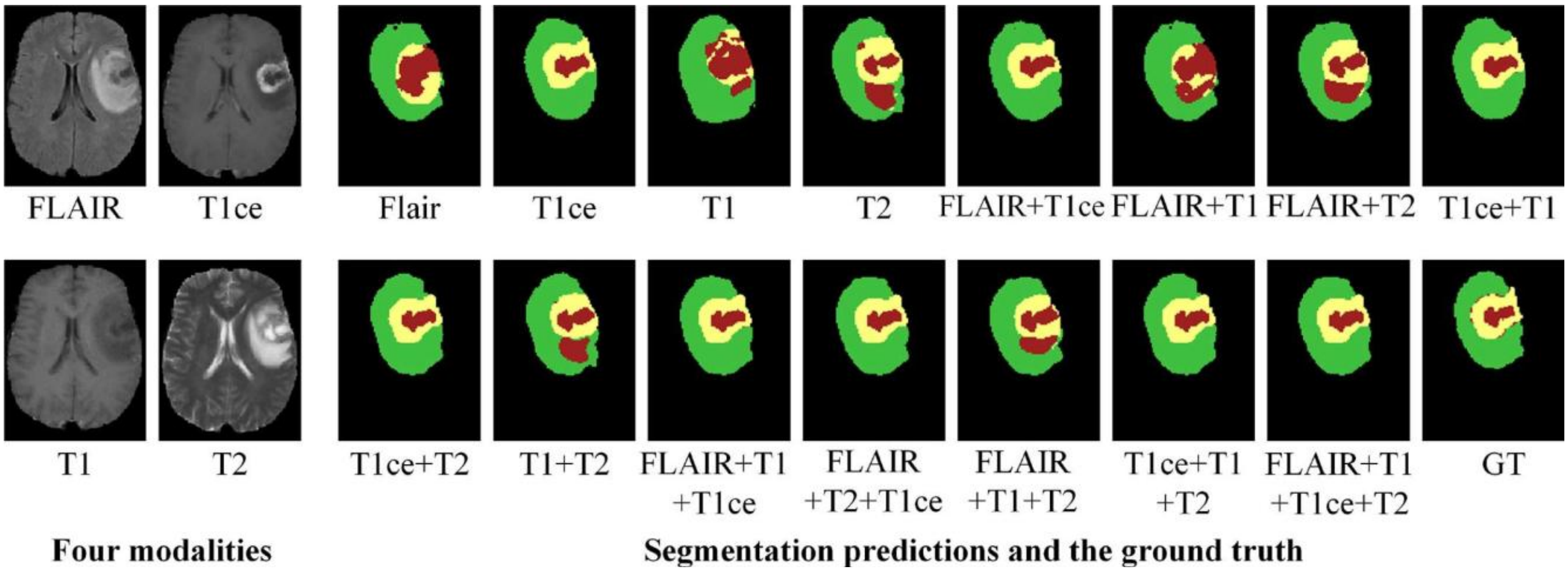

Fig. 5. Qualitative comparison for segmentation results of fifteen different missing modalities. Four different modalities are on the left, and segmentation results and ground truth of the proposed model are in fifteen different combinations on the right.

TABLE VII
ABLATION STUDY ON THE EFFECTIVENESS OF THE MAMBA BLOCK

| Method | Average bAHD (mm) ↓ | | |
|---|---|---|---|
| | WT | TC | ET |
| W/O UFE | 7.20 | 8.64 | 6.26 |
| W/O Mamba in UFE | 6.19 | 7.27 | 5.56 |
| W/O CMF | 8.54 | 9.27 | 7.43 |
| W/O Mamba in CMF | 6.10 | 7.03 | 5.82 |
| W/O MHSA in UFE & CMF | 7.33 | 8.17 | 6.09 |
| W/O Mamba in UFE & CMF | 6.37 | 7.85 | 5.91 |
| MMTSeg | 4.14 | 6.24 | 3.75 |

Note: W/O Mamba in UFE or CMF represent removing Mamba from either UFE or CMF. W/O MHSA in UFE & CMF and W/O Mamba in UFE & CMF represent removing MHSA or Mamba from both UFE and CMF to form a pure Mamba network or a pure Transformer network. All the methods are trained under the same protocol as the proposed model.

TABLE VIII
ABLATION STUDY OF THE EFFECTIVENESS OF THE CMF

| Method | Average bAHD (mm) ↓ | | |
|---|---|---|---|
| | WT | TC | ET |
| Convolutional Fusion | 10.54 | 11.27 | 8.43 |
| Transformer Fusion | 4.84 | 6.97 | 4.13 |
| MMTSeg | 4.14 | 6.24 | 3.75 |

Note: Convolutional Fusion and Transformer Fusion represent replacing CMF by a convolutional and Transformer fusion module, respectively.

and Transformer-based fusion modules, respectively. The results shown in Table VIII demonstrate that our proposed hybrid Mamba-Tranformer CMF is more effective for incomplete brain tumor segmentation compared to the 3-D convolution fusion decoder (i.e., Convolutional Fusion) in RFNet and the modality-correlated encoder (i.e., Transformer Fusion) in mmFormer.

## *F. Qualitative Analysis*

In Fig. 4, we visualize our model, mmFormer, and RobustSeg in full modalities, missing one modality (FLAIR + T1 + T2), missing two modalities (T1 + T2), and missing three modalities. The segmentation results in state (T1), and the visualization results show that in most cases, our method can obtain accurate segmentation results. At the same time, we further analyzed the impact of different modes of missing data. It can be seen from Fig. 4 that the T1 modality plays an important role in the segmentation of the entire tumor. Even if the other three modalities are missing, the T1 modality can obtain more accurate segmentation results. The region is the most sensitive, and when the T1ce modality is missing, the segmentation effect of gangrene and nonenhancing tumor regions is severely degraded, resulting in inaccurate boundary segmentation. When the FLAIR modality is missing, the segmentation accuracy of peritumoral edema decreases.

In Fig. 5, we show the segmentation results of our model in the absence of 15 different modalities. It is evident that our proposed method can adapt to the absence of different modalities and can obtain accurate segmentation results even with only one modality. We can draw a conclusion from the qualitative comparisons of segmentation prediction results and the ground truth. First, it is obvious that the segmentation predictions are more similar to the ground truth when there are more modalities input into our model. Second, the T1ce modality in MRI is more helpful for our model predictions because the predictions with the T1ce modality are better than those of other modalities' missing conditions. Third, our model learned less useful information from the modality T1 to make a good prediction, due to the worst predictions with T1 shown in the figure. This indicates that the performance of our model may be further boosted if it can learn more from the modality, which is an improvement direction in future research.

Distinct from other state-of-the-art models, the proposed MMSeg has a Mamba-Transformer CMF module instead of pure CNN or Transformer-based fusion. Our proposed CMF module aims to extract modality-invariant feature representations by establishing and aligning global correlations between modalities. It leverages the attention mechanism of the Transformer to focus on relevant modality-specific features corresponding to tumor regions, compensating for the absence of certain modalities. The comparison results of the qualitative analysis show that different modalities have different sensitivities to different tumor regions, and the CMF module we designed can make full use of the complementary characteristics of different modalities to improve the overall performance of the model.

## V. Conclusion

We developed a novel model that utilizes feature distillation to guide the segmentation of incomplete multimodal brain tumors, resulting in high-precision tumor segmentation. With the support of feature distillation, our approach leverages complementary knowledge of complete modal MRI to guide the feature enhancement process. Complementary information distilled across multimodalities and modality-invariant feature representations extracted from each unimodality are effectively enhanced and fused via novel Mamba-Transformer hybrid networks to determine missing modalities. The experimental evaluation and ablation study demonstrate that the proposed Mamba-Transformer hybrid networks play a key role in improving performance on brain tumor segmentation with incomplete MRI modalities by modeling the semantic relationship with long-range dependencies dynamically. The boundary-aware loss function ensures that the limited information available is used to accurately define the tumor edge when the boundary-specific modules with Mamba-Transformer hybrid attention mechanisms are integrated, capturing fine-grained details even when some modal data is unavailable.

The methodological implications discussed above rely on both the Mamba's use of continuous dynamics for sequence representation and the Transformer encoder architecture to capture dependencies between tokens within a sequence, in the form of different unimodal compact features extracted by CNN encoders, which learn a 3-D representation of an input MRI image. However, the proposed MMTSeg is limited to brain tumor MRI segmentation and the unique architecture of the hybrid Mamba and Transformer for feature enhancement and fusion. Due to Mamba-Transformer being an emerging framework, and Mamma in computer vision still being in development, we investigated only one hybrid Mamba-Transformer architecture, rather than the others, aiming for fewer computational requirements.

In future research work, the proposed MMTSeg should be extended to consider other abdominal organs or medical imaging modalities in semantic segmentation with missing modalities, such as liver computed tomography (CT) segmentation or combined MRI and CT segmentation in multiorgan settings. Such an extension might potentially require combining the specific modality with other techniques for multimodal feature extraction and CMF, for example, missing modality compensation via real and synthetic images using diffusion models. Furthermore, we would have liked to conduct a postanalysis to evaluate the benefits of the Mamba block with Transformer across different architectures.